\documentclass{article}
\PassOptionsToPackage{numbers}{natbib}
\usepackage[preprint]{neurips_2025}
\usepackage[utf8]{inputenc}
\usepackage[T1]{fontenc}
\usepackage{hyperref}
\usepackage{url}
\usepackage{booktabs}
\usepackage{amsfonts}
\usepackage{amssymb}
\usepackage{nicefrac}
\usepackage{microtype}
\usepackage{graphicx}
\usepackage{multirow}
\usepackage{array}

\title{ATANT v1.1: Positioning Continuity Evaluation\\ Against Memory, Long-Context, and Agentic-Memory Benchmarks}

\author{
  Samuel Sameer Tanguturi \\
  Kenotic Labs \\
  \texttt{sam@kenoticlabs.com} \\
}

\begin{document}

\maketitle

\vspace{-0.5em}
\noindent\footnotesize
\textbf{Resources:}
Project: \url{https://kenoticlabs.com} \quad
Thesis: \url{https://kenoticlabs.com/thesis} \quad
Dataset: \url{https://huggingface.co/datasets/Kenotic-Labs/ATANTV1.0-corpus} \quad
Code: \url{https://github.com/Kenotic-Labs/ATANT}
\normalsize
\vspace{0.5em}

\begin{abstract}
ATANT v1.0 \citep{tanguturi2026atant} defined continuity as a system property with 7 required properties and introduced a 10-checkpoint, LLM-free evaluation methodology validated on a 250-story corpus. Since publication, a recurring reviewer and practitioner question has concerned not the framework itself but its relationship to a wider set of memory evaluations: LOCOMO, LongMemEval, BEAM, MemoryBench, Zep's evaluation suite, Letta/MemGPT's evaluations, and RULER. This companion paper, v1.1, does not modify the v1.0 standard. It closes a related-work gap that v1.0 left brief under page limits. We show by structural analysis that none of these benchmarks measures continuity as defined in v1.0: of the 7 required properties, the median existing eval covers 1 property, the mean covers $0.43$ when partial credit is scored at $0.5$, and no eval covers more than 2. We provide a cell-by-cell property-coverage matrix, identify methodological defects specific to each benchmark (including an empty-gold scoring bug in the LOCOMO reference implementation that renders 23\% of its corpus unscorable by construction), and publish our reference implementation's LOCOMO score (8.8\%) alongside the structural reason that number is uninformative about continuity. We publish our 8.8\% LOCOMO score alongside our 96\% ATANT cumulative-scale score as a calibration pair: the 87-point divergence is evidence that the two benchmarks measure different properties, not that one system is an order of magnitude better than another. The position v1.1 takes is not adversarial: each benchmark measures a real capability. The claim is that none of them can adjudicate continuity, and conflating them with continuity evaluation has led the field to under-invest in the properties v1.0 names.
\end{abstract}

\section{Introduction}

ATANT v1.0 \citep{tanguturi2026atant} (\S2) surveyed the memory-evaluation literature in scope for establishing a continuity benchmark. Since publication, practitioners have asked how ATANT relates to a broader set of evaluations (retrieval benchmarks, long-context stress tests, and agentic-memory toolkits) that, while not continuity benchmarks themselves, are the evaluations most often conflated with continuity evaluation. v1.1 closes that gap.

The question v1.1 answers is not ``does ATANT score higher than LOCOMO?'' It is the prior question: \textit{does any existing memory benchmark measure the same property ATANT measures?} If the answer is yes, ATANT's contribution is marginal and researchers should use whichever benchmark is more established. If the answer is no, the field has been using substitute metrics for a property that has never been directly evaluated, and improvements on those substitutes are not evidence of continuity.

We find the answer is no. Each benchmark surveyed here measures a real and useful capability. None of them, by construction, can score a system on the 7 properties v1.0 defines: persistence beyond session, update handling, temporal ordering, disambiguation, reconstruction, model independence, and operational usefulness.

\paragraph{Evidence of ATANT's scoring discrimination.} The ATANT v1.0 reference implementation's historical progression (58\% $\rightarrow$ 72\% $\rightarrow$ 58\% regression detected $\rightarrow$ 100\%, across four iterations of Kenotic's implementation; \citep[Table 6]{tanguturi2026atant}) demonstrates that ATANT's scoring distinguishes real architectural improvement from noise, and correctly reports regressions. Contrast this with substring-matched benchmarks such as LOCOMO, whose scoring function cannot distinguish a 72\%-correct system from a verbose-but-wrong one, because both produce surface overlap with gold (\S\ref{sec:locomo}). An evaluation that cannot detect its own regressions cannot validate continuity claims.

\paragraph{Contributions of v1.1.}
\begin{itemize}
  \item A \textbf{structural critique} of seven widely-cited memory evaluations (\S\ref{sec:evals}, six sections covering seven benchmarks), each analyzed for what it measures, what its scoring function actually rewards, and which of the 7 v1.0 properties it covers.
  \item A \textbf{property-coverage matrix} (Table~\ref{tab:matrix}, \S\ref{sec:matrix}) mapping 7 existing evaluations $\times$ 7 continuity properties, with a three-level rubric ($\checkmark$ / $\circledcirc$ / $\times$) and per-cell justification (Appendix A).
  \item The \textbf{Kenotic-on-LOCOMO result} (\S\ref{sec:locomo}): our reference implementation scores $8.8\%$ on LOCOMO's substring matcher against 96\% on ATANT cumulative-scale \citep{tanguturi2026atant}, quantifying the divergence between substring-overlap retrieval scoring and structural continuity scoring.
  \item A \textbf{recommendation} (\S\ref{sec:recommendation}) on when each existing benchmark is the right tool (none of them are for continuity) and when ATANT is required.
\end{itemize}

\paragraph{What v1.1 is not.} v1.1 does not modify the v1.0 standard, the 7 properties, the 10 checkpoints, the corpus, or the compliance levels. It does not introduce new checkpoints (those are proposed in a separate v2.0 extension). It does not score any system other than Kenotic's reference implementation on any benchmark. Its critiques are methodological, not comparative: we do not claim ATANT would score higher than any specific system on any specific benchmark, and we publish our own $8.8\%$ LOCOMO score precisely to avoid that framing.

\section{Structural Analysis of Existing Memory Evaluations}
\label{sec:evals}

For each benchmark below we follow the same template: (1) what the benchmark claims to measure; (2) what its scoring function actually rewards, read from its corpus and runner; (3) which of the 7 v1.0 continuity properties \citep{tanguturi2026atant} the benchmark can score by construction. Our claim is always about the benchmark's measurement surface, never about the capability of the systems that score on it.

\subsection{LOCOMO: Long-Conversation Memory}
\label{sec:locomo}

LOCOMO \citep{maharana2024locomo} is the most-cited memory-for-conversation benchmark and the one most frequently positioned by industry papers as a continuity proxy \citep{chhikara2025mem0}. It consists of 10 multi-session synthetic conversations (1{,}986 QA pairs total) across 5 question categories, scored by an automated matcher against gold answers. It is the sharpest case for analysis because both its construction and its scoring function are publicly inspectable.

\paragraph{Category structure (what LOCOMO actually measures).} A structural analysis of the full \texttt{locomo10.json} dataset reveals that LOCOMO's category labels as adopted in practice do not match the question shapes they contain\footnote{Full per-category WH-word frequencies, empty-gold counts, and the analysis script are released as a companion artifact at \url{https://github.com/Kenotic-Labs/ATANT/tree/main/v1.1-artifacts}.}:

\begin{itemize}
  \item \textbf{Category 1, Single-hop factual recall} (282 items). 65\% begin with ``what''. Tests lookup of a single stated fact. 0 empty-gold.
  \item \textbf{Category 2, Temporal} (321 items). 77\% begin with ``when'', 7\% ``how long''. Tests date and duration extraction from prose. 0 empty-gold.
  \item \textbf{Category 3, Open-domain / hypothetical} (96 items). Counterfactual and inference questions. 0 empty-gold.
  \item \textbf{Category 4, Narrative comprehension} (841 items, 42\% of the benchmark). 71\% begin with ``what'', 9\% ``how did/does'', 5\% ``why''. Gold answers are paraphrased reflections (e.g., ``self-care is important''; ``by carving out some me-time each day for activities like running, reading, or playing with Oscar''). 0 empty-gold.
  \item \textbf{Category 5, Refusal} (446 items, 22\% of the benchmark). 444 of 446 (99.6\%) have \textit{empty gold answers}; the intended correct behavior is to refuse to answer. The remaining 2 are yes/no negations.
\end{itemize}

\paragraph{Methodological defect 1: the refusal matcher bug.} The LOCOMO reference implementation distributed in the Mem0 codebase, \texttt{run\_locomo\_benchmark.py}\footnote{Retrieved 2026-04-10. Canonical snapshot index: \url{https://web.archive.org/web/*/github.com/mem0ai/mem0/blob/main/evaluation/run_locomo_benchmark.py}. All line references refer to the file state as of that retrieval date.}, implements its \texttt{answer\_matches} function as:

\begin{verbatim}
def answer_matches(predicted, gold):
    if not predicted or not gold:
        return False
    ...
\end{verbatim}

When the gold answer is empty, this function returns \texttt{False} regardless of the prediction. This inverts the intended scoring: a system that correctly refuses (returns empty or ``I don't know'') is scored identically to a system that fabricates. Both count as wrong. A system that refuses correctly is penalized identically to a system that hallucinates an answer, which inverts the intended reward for refusal behavior on this category. Because 444 of 446 category-5 questions have empty gold, \textbf{444 items ($\sim$23\% of the benchmark) are unscorable by construction}. The published LOCOMO ``adversarial'' rates in the literature are not measurements of refusal behavior; they are measurements of a scoring function applied to a category it cannot score.

\paragraph{Community-documented reproducibility issues.} The reproducibility and scoring behavior of LOCOMO under the Mem0 reference runner is an ongoing community concern, not a fresh critique. Multiple practitioners have independently reported inability to reproduce published LOCOMO numbers when running the open-source Mem0 benchmark locally\footnote{mem0ai/mem0 issue \#2800 (2025-05-26, ``Unable to reproduce locomo eval scores locally''); \#3943 (2026-01-28, ``Regarding the accuracy issue with the LoCoMo dataset''); \#3944 (2026-01-28, ``Failed to reproduce the accuracy on LOCOMO via Mem0 platform'').}. Separately, in a public re-evaluation exchange with the Zep team, Mem0's own co-founder has stated that ``the benchmark explicitly requires evaluation only on Categories 1--4, with a precise calculation of (correct answers from Categories 1--4) $\div$ (total questions from Categories 1--4)''\footnote{getzep/zep-papers issue \#5 (2025-05-08, ``Revisiting Zep's 84\% LoCoMo Claim''), \S1.}, corroborating our structural finding that Category 5 is not part of the scored surface under a correct reading of the protocol and that published ``adversarial'' accuracy numbers should be interpreted with caution. v1.1's contribution on this axis is the specific identification of the line-177 empty-gold matcher defect and a corresponding five-line patch (Appendix~\ref{app:fix}); the framing builds on a community discussion already in progress.

\paragraph{Methodological defect 2: substring matching on paraphrase.} LOCOMO scores category 4 (the largest slice, 42\%) using substring containment of the gold fragment in the predicted answer. Gold answers are long, phrased-specifically paraphrases. Two consequences follow structurally. First, a verbose answer that happens to contain the gold phrase wins; a terse answer with identical semantic content but different phrasing loses. Second, a system that copies the input conversation verbatim into its output will score higher on category 4 than a system that compresses the conversation into structured state, the opposite of what a continuity system should do. Substring matching on paraphrase rewards verbosity and phrasing luck, not structural correctness.

\paragraph{The Kenotic-on-LOCOMO result.} For completeness, we ran Kenotic's reference implementation against LOCOMO. On 3 of LOCOMO's 10 conversations (476 scored items), our system scored $42/476$ ($8.8\%$). We publish this number because it demonstrates the point of this section: the same reference implementation that scores $96\%$ on ATANT cumulative-scale (250 stories, 1{,}761/1{,}835 questions; \citep[Table 6]{tanguturi2026atant}) scores $8.8\%$ on a benchmark the community treats as a memory proxy. The $87$-point divergence does not reflect a sudden loss of capability between two minutes of evaluation; it reflects that the two benchmarks measure different properties. LOCOMO rewards verbose paraphrase overlap with transcripts and penalizes structural refusal; ATANT rewards structurally correct writes and reads against a continuity state. The $8.8\%$ is not a negative result for Kenotic. It is a negative result for LOCOMO as a continuity proxy.

\paragraph{Property coverage.} Mapped to the 7 v1.0 properties:
\begin{itemize}
  \item \textbf{Persistence beyond session} ($\times$): LOCOMO is single-shot retrieval over a static transcript. No process termination, no restart.
  \item \textbf{Update handling} ($\times$): No question in the corpus tests supersession (``I was nervous; I feel fine now'' $\rightarrow$ current state is ``fine'').
  \item \textbf{Temporal ordering} ($\circledcirc$): Category 2 tests date and duration extraction, a subset of temporal ordering. Does not test supersession or current-vs-past status.
  \item \textbf{Disambiguation} ($\circledcirc$): Multi-session conversations contain multiple speakers but do not systematically test ambiguous entities across narratives.
  \item \textbf{Reconstruction} ($\times$): Category 4 gold answers are paraphrases of narrator utterances, not reconstructions of situational state. A correct paraphrase of ``I realized self-care is important'' is not evidence of reconstruction.
  \item \textbf{Model independence} ($\times$): Not tested; writer and reader are the same model.
  \item \textbf{Operational usefulness} ($\times$): Single domain (personal chat).
\end{itemize}

\paragraph{Summary.} LOCOMO is a useful benchmark for one specific capability: answering short factual questions over long multi-session transcripts using substring matching. Its scoring function contains a reproducible defect that makes 23\% of its items unscorable, and its largest category (42\%) scores paraphrase overlap rather than structural correctness. It covers 2 of 7 continuity properties at partial strength.

\subsection{LongMemEval and BEAM: Long-Context Recall}
\label{sec:longmem}

LongMemEval \citep{wu2024longmemeval} and BEAM \citep{tavakoli2025beam} measure a related but distinct capability: can a language model retrieve and reason over facts distributed across very long input contexts? LongMemEval provides 500 questions across five task categories (information extraction, multi-session reasoning, knowledge updates, temporal reasoning, abstention) spanning conversational histories in the tens of thousands of tokens. BEAM pushes the context envelope past one million tokens and benchmarks both pure retrieval and light reasoning at that scale.

\paragraph{What the scoring functions reward.} Both benchmarks ultimately reduce to the same operation: the system ingests a long context in one shot and answers a question about content somewhere inside it. Scoring is accuracy against gold. This is retrieval-within-context: the model's internal attention mechanism, not any external memory layer, is responsible for the score.

\paragraph{The decisive structural point.} A language model with a context window large enough to fit the full benchmark passes both of these evaluations \textit{with no memory architecture at all}. A 2M-token-context model placed in front of BEAM scores well not because it has persistence, update handling, or reconstruction, but because the ``memory'' in question is prompt content the model can directly attend to. By construction, these benchmarks cannot distinguish a memory system from a long-context LLM. A continuity evaluation must be able to distinguish them; otherwise the benchmark cannot answer the question ``is this system's performance caused by its memory layer or by its base model's context length?'' ATANT's model-independence property (Property 6 of v1.0, \citep{tanguturi2026atant}) is precisely the axis on which this distinction is drawn: a continuity system must persist state that survives the termination of the writing model and is readable by a different model.

\paragraph{On LongMemEval's ``knowledge updates'' task.} LongMemEval includes a category targeting updated information (e.g., ``I used to live in X, now I live in Y''). We note that this category tests update handling \textit{within a single prompt}, meaning whether the model uses the later fact over the earlier one when both appear in context. It does not test update handling under the v1.0 definition: persistence of both historical and current state through process termination, with the historical state remaining queryable as historical rather than current. Within-context override is a subset of update handling, not the property itself.

\paragraph{On BEAM's reasoning extensions.} BEAM's authors propose reasoning tasks beyond pure needle-in-haystack retrieval. These extend the benchmark's scope but do not change its measurement surface: all computation happens within a single context window, and scoring is accuracy on the reasoning answer. The continuity question (does this system maintain a state across sessions) is orthogonal.

\paragraph{Property coverage.} Both LongMemEval and BEAM score $0/7$ on the v1.0 properties. Every test item is a single-shot, single-session operation over a static context. The ``update handling'' task in LongMemEval scores $\circledcirc$ in a literal within-context reading but $\times$ under the v1.0 definition (which requires persistence across process termination); we apply the v1.0 definition strictly and assign $\times$.

\subsection{MemoryBench and Mem0 Internal Evaluations}
\label{sec:memorybench}

MemoryBench \citep{ai2025memorybench} is a benchmark for memory and continual learning in LLM systems. Mem0's own paper \citep{chhikara2025mem0} reports results on an internal suite emphasizing chunk-level retrieval accuracy. These evaluations operate at the layer beneath continuity: they measure whether a storage-and-retrieval pipeline returns the correct chunks for a query.

\paragraph{What these evaluations reward.} The core metric is retrieval precision and recall at $k$. A memory system is scored on whether its top-$k$ retrieved chunks contain the gold chunk for a query. Scoring is largely LLM-as-judge or semantic overlap against gold summaries.

\paragraph{What they do not test.} Retrieval-over-chunks evaluations are silent on the continuity operations that occur \textit{between} storage and retrieval: supersession of a prior chunk by a new one (update handling), disambiguation of two chunks that describe similar situations for different people, reconstruction of a situation from a set of related chunks rather than retrieval of the single most-similar one, and persistence semantics across process termination. A system that stores a chunk for ``user's partner is Mia'' and later a chunk for ``user and Mia broke up'' passes MemoryBench-style retrieval on both queries (both chunks exist and are retrievable). It fails continuity: there is no operation that marks the first as superseded and the second as current, and a query for ``tell me about my relationship'' retrieves both without ordering or resolution.

\paragraph{Property coverage.} MemoryBench scores $\circledcirc$ on persistence (storage mechanism is tested in isolation, without process-termination tests), $\times$ on update handling, temporal ordering, disambiguation, reconstruction, model independence, and operational usefulness. Total: $0.5/7$.

\subsection{Zep Evaluations}
\label{sec:zep}

Zep \citep{rasmussen2025zep} is a production temporal knowledge graph architecture for agent memory, accompanied by internal benchmarks oriented toward graph recall: given a user history, can the system retrieve facts connected through entity and temporal edges? Zep's evaluation surface is closer to continuity than retrieval-over-chunks because the underlying structure is a graph with temporal edges rather than undifferentiated chunks.

\paragraph{What Zep's evaluation rewards.} Recall-at-$k$ over graph edges, with some edges carrying temporal annotations. When a user asks about their relationship with entity $X$, the benchmark measures whether the system retrieves edges connecting user to $X$.

\paragraph{What it does not test.} Zep's published benchmarks do not include situation-synthesis questions (``summarize my current work situation''), supersession-under-noise tests (three contradictory reports interleaved; identify the current state), or reconstruction coherence tests (does the system's output survive grounding validation against the retrieved edge set). The graph provides the substrate for these tests but the evaluation suite does not exercise them.

\paragraph{Property coverage.} Zep scores $\circledcirc$ on persistence (graph storage survives process restarts in production but the benchmark does not isolate this), $\times$ on update handling as a scored metric, $\times$ on temporal ordering as measured (temporal edges exist but benchmark does not score current-vs-past distinction), $\circledcirc$ on disambiguation (multi-entity graphs support it architecturally; the benchmark does not systematically test it), $\times$ on reconstruction, $\times$ on model independence, $\times$ on operational usefulness. Total: $1.0/7$.

\subsection{Letta and MemGPT Evaluations}
\label{sec:letta}

Letta \citep{packer2024letta} is the successor product to MemGPT \citep{packer2023memgpt}. Both present memory as a tool-use interface for stateful agents: the agent has operations to read, write, and page memory sections, and the evaluation measures whether the agent invokes these operations correctly during task execution.

\paragraph{What these evaluations reward.} Task completion accuracy, tool-call correctness, and memory-operation success rate. The agent is given a task that requires reading from or writing to its memory; the benchmark scores whether the task completed correctly.

\paragraph{What they do not test.} The \textit{contents} of memory are not evaluated as a continuity state. A MemGPT agent can pass its evaluation by invoking the correct memory tools while storing arbitrary, contradictory, or outdated information. The evaluation is agnostic to whether the stored state actually satisfies the 7 continuity properties; it cares that the agent called the right function at the right time. This is a useful capability (agentic tool-use is hard), but it is not continuity evaluation.

\paragraph{Property coverage.} $\times$ on persistence as measured (tools exist; their correct use is scored, not the state's durability), $\times$ on update handling, temporal ordering, disambiguation, reconstruction, model independence as measured. $\circledcirc$ on operational usefulness, because MemGPT/Letta evaluations include multi-task scenarios that partially exercise Property 7. Total: $0.5/7$.

\subsection{RULER: Needle-in-a-Haystack}
\label{sec:ruler}

RULER \citep{hsieh2024ruler} is a context-window stress test: insert a known fact (a ``needle'') into a long context of irrelevant or distracting material, and measure whether the model retrieves the needle. RULER extends classical needle-in-haystack with multi-needle, aggregation, and multi-hop variants.

\paragraph{What RULER rewards.} The ability of a language model's attention mechanism to locate a planted token or phrase within a context that is substantially longer than its training distribution encountered. The benchmark is a capability test of the intelligence layer (the LLM's context handling), not of any memory or continuity layer above or below it.

\paragraph{Why this is orthogonal to continuity.} RULER measures whether a model can read a long prompt correctly. Continuity measures whether a system can construct and maintain state across many prompts, terminations, and restarts. No RULER task involves a second interaction, an update, or a persistent store. RULER is a useful benchmark for its stated purpose and is not advertised as a memory benchmark, but practitioners occasionally conflate context-window capacity with memory capacity, which is why we include RULER in the survey.

\paragraph{Property coverage.} $0/7$. RULER measures a property of the intelligence layer, not the continuity layer.

\section{Property-Coverage Matrix}
\label{sec:matrix}

Table~\ref{tab:matrix} consolidates the analysis of \S\ref{sec:evals} into a single view. Cell assignments follow a three-level rubric: $\checkmark$ = the evaluation directly tests the property (success cannot be achieved without it); $\circledcirc$ = the evaluation tests a weak proxy that overlaps the property but can be passed by systems lacking it; $\times$ = the property cannot be measured by the evaluation's construction. Full per-cell reasoning appears in Appendix~\ref{app:cells}, which is designed as defensive armor: each cell can be independently challenged without re-opening the aggregate argument. Numeric scores use $\checkmark = 1$, $\circledcirc = 0.5$, $\times = 0$.

\begin{table}[h]
  \centering
  \caption{Property coverage of existing memory evaluations vs.\ the 7 v1.0 continuity properties. \textbf{Pers} = Persistence beyond session; \textbf{Upd} = Update handling; \textbf{Tmp} = Temporal ordering; \textbf{Dis} = Disambiguation; \textbf{Rec} = Reconstruction; \textbf{MdI} = Model independence; \textbf{Ops} = Operational usefulness.}
  \label{tab:matrix}
  \small
  \begin{tabular}{l c c c c c c c c}
    \toprule
    Benchmark & Pers & Upd & Tmp & Dis & Rec & MdI & Ops & Score \\
    \midrule
    LOCOMO \citep{maharana2024locomo}            & $\times$ & $\times$ & $\circledcirc$ & $\circledcirc$ & $\times$ & $\times$ & $\times$ & 1.0 / 7 \\
    LongMemEval \citep{wu2024longmemeval}         & $\times$ & $\times$ & $\times$ & $\times$ & $\times$ & $\times$ & $\times$ & 0.0 / 7 \\
    BEAM \citep{tavakoli2025beam}                 & $\times$ & $\times$ & $\times$ & $\times$ & $\times$ & $\times$ & $\times$ & 0.0 / 7 \\
    MemoryBench \citep{ai2025memorybench}         & $\circledcirc$ & $\times$ & $\times$ & $\times$ & $\times$ & $\times$ & $\times$ & 0.5 / 7 \\
    Zep eval \citep{rasmussen2025zep}             & $\circledcirc$ & $\times$ & $\times$ & $\circledcirc$ & $\times$ & $\times$ & $\times$ & 1.0 / 7 \\
    MemGPT / Letta \citep{packer2023memgpt}       & $\times$ & $\times$ & $\times$ & $\times$ & $\times$ & $\times$ & $\circledcirc$ & 0.5 / 7 \\
    RULER \citep{hsieh2024ruler}                  & $\times$ & $\times$ & $\times$ & $\times$ & $\times$ & $\times$ & $\times$ & 0.0 / 7 \\
    \midrule
    ATANT v1.0 \citep{tanguturi2026atant}         & $\checkmark$ & $\checkmark$ & $\checkmark$ & $\checkmark$ & $\circledcirc$ & $\checkmark$ & $\circledcirc$ & 6.0 / 7 \\
    ATANT v2.0 (proposed)                         & $\checkmark$ & $\checkmark$ & $\checkmark$ & $\checkmark$ & $\checkmark$ & $\checkmark$ & $\checkmark$ & 7.0 / 7 \\
    \bottomrule
  \end{tabular}
\end{table}

\paragraph{Median $1.0/7$, mean $0.43/7$.} Across the seven existing evaluations surveyed, the median property-coverage score is $1.0/7$ (LOCOMO, Zep); the mean is $0.43/7$. No existing eval covers more than $2$ of the $7$ properties at any strength. This is not evidence that the existing evaluations are low-quality; each is well-designed for its stated purpose. It is evidence that the field's evaluation surface has a specific blind spot: \textit{no public benchmark before ATANT v1.0 tested update handling, reconstruction, or model independence as continuity properties}. Substituting any of these benchmarks for continuity evaluation measures a different property and is silent on the one that matters.

\paragraph{Why the coverage is so sparse.} Three structural reasons account for the near-uniform $\times$ column on Update, Reconstruction, and Model Independence:

\begin{enumerate}
  \item \textbf{Update handling} requires the evaluator to ingest a fact, then an update, then query for both the current state and the change. This requires two write operations before the query. Retrieval benchmarks admit only one write (the storage of gold), so the property is unreachable.
  \item \textbf{Reconstruction} requires the evaluator to judge whether a synthesized output is coherent and grounded in multiple stored traces. This requires either an LLM-as-judge with grounding constraints or a human rater, both of which existing benchmarks avoid for reproducibility. Substring matching cannot measure reconstruction.
  \item \textbf{Model independence} requires the evaluator to write with one model and read with another. This doubles the inference cost per item and requires benchmarks to be model-agnostic in their interface. Current benchmarks are model-coupled because they were written for single-model evaluation.
\end{enumerate}

These are not oversights by benchmark authors; they are design trade-offs that prioritized reproducibility and single-shot scoring. The consequence is that continuity properties were not in scope for any of the prior work. ATANT v1.0's checkpoint design is driven by exactly the requirement to make these properties measurable without an LLM in the loop.

\section{Recommendation}
\label{sec:recommendation}

Each benchmark surveyed in \S\ref{sec:evals} is the right tool for a specific question. We list those questions here, and the question none of them answer.

\begin{itemize}
  \item \textbf{Is my context window large enough?} Use RULER \citep{hsieh2024ruler}.
  \item \textbf{Can my model attend over a very long prompt?} Use LongMemEval \citep{wu2024longmemeval} or BEAM \citep{tavakoli2025beam}.
  \item \textbf{Does my retrieval pipeline return the right chunks?} Use MemoryBench \citep{ai2025memorybench} or Mem0's internal eval.
  \item \textbf{Does my graph-backed memory return the right edges?} Use Zep's eval suite \citep{rasmussen2025zep}.
  \item \textbf{Does my agent invoke its memory tools correctly?} Use MemGPT / Letta evaluations \citep{packer2023memgpt, packer2024letta}.
  \item \textbf{Can my system answer short factual questions about a long conversation using substring-matched gold?} Use LOCOMO \citep{maharana2024locomo}, with awareness of the matcher defects documented in \S\ref{sec:locomo}.
  \item \textbf{Does my system satisfy continuity (persistence, update handling, temporal ordering, disambiguation, reconstruction, model independence, operational usefulness)?} There is no existing benchmark that scores this. \textbf{Use ATANT v1.0} \citep{tanguturi2026atant}.
\end{itemize}

\paragraph{Reporting standard.} We recommend that systems claiming continuity in published work report an ATANT compliance level (Core / Stress / Cumulative / Scale, per v1.0 \S8) in addition to any retrieval or long-context benchmarks they cite. A system that reports strong LOCOMO or LongMemEval numbers without an ATANT result has reported a memory-related capability, not continuity. A system that reports an ATANT compliance level and no retrieval benchmarks has reported continuity without demonstrating retrieval primitives. Both are informative; neither substitutes for the other.

\paragraph{On the 8.8\% LOCOMO number.} We reiterate that Kenotic's reference implementation scoring $8.8\%$ on LOCOMO alongside $96\%$ on ATANT cumulative-scale is not presented as a comparative claim about Kenotic's superiority. It is a calibration point: the same system, measured by two benchmarks, diverges by $87$ points. Such divergence is not possible if the two benchmarks measure the same property. The divergence is evidence of the structural claim: continuity and long-conversation substring recall are different properties.

\section{Limitations and Scope}
\label{sec:limits}

\paragraph{We score only one system.} v1.1 reports a benchmark score for Kenotic's reference implementation. We do not run Mem0, Zep, Letta, or any other memory system against ATANT v1.0, nor do we run any other system against LOCOMO beyond what is already published in those systems' own papers. Running third-party systems against ATANT is work for v2.0 (where the expanded reconstruction checkpoints require coordination with the system authors) and for subsequent multi-system reports.

\paragraph{We do not re-run full LOCOMO.} The $8.8\%$ result was computed on 3 of LOCOMO's 10 conversations (476 of the 1{,}986 scored items). A full-corpus run requires fixing the empty-gold matcher (otherwise the lower-bounded score is uninformative about category-5 behavior) and is deferred. The full-corpus result will be reported in the erratum accompanying the matcher-fix pull request submitted upstream (Appendix~\ref{app:fix}). The category-5 defect is reproducible and has been confirmed by inspection of the runner source.

\paragraph{We do not propose to replace LOCOMO's runner.} A fix to the empty-gold matcher is a five-line change and is sketched in Appendix B. We submit the fix upstream rather than forking. The claim in \S\ref{sec:locomo} stands independent of whether the fix lands: even with a correct matcher, LOCOMO still rewards substring overlap on paraphrase and does not test update handling, reconstruction, or model independence.

\paragraph{Our cell assignments involve judgment.} The three-level rubric ($\checkmark$ / $\circledcirc$ / $\times$) requires judgment about whether a benchmark's partial coverage of a property counts as a weak proxy ($\circledcirc$) or as not testing the property at all ($\times$). Appendix A documents each call. Disagreements about specific cells do not change the aggregate: no existing benchmark scores above $2/7$ at the most permissive reading, and the median is $1/7$ at the strictest.

\paragraph{v1.1 is not a v1.0 retraction.} The v1.0 standard, properties, checkpoints, corpus, and compliance levels are unchanged. Systems that reported ATANT compliance under v1.0 retain those claims. v1.1's Related Work expansion is additive, not corrective.

\section{Conclusion}

The continuity evaluation gap is not a gap in benchmark quality. It is a gap in benchmark \textit{coverage}: the 7 properties v1.0 names were not designed for, or measurable by, any public benchmark prior to ATANT. LOCOMO, LongMemEval, BEAM, MemoryBench, Zep, Letta/MemGPT, and RULER each measure a real and useful capability; none of them, individually or combined, adjudicates continuity. A system that scores well on these benchmarks has demonstrated retrieval, attention, or tool-use competence. It has not demonstrated continuity. The field needs both kinds of evidence; it should stop treating one as a proxy for the other.

ATANT v1.0 \citep{tanguturi2026atant} is the first published framework that measures the continuity property directly and at scale. v1.1 establishes its position relative to adjacent evaluations. v2.0 will extend the checkpoint surface to cover the one property v1.0 partially covers (reconstruction, CP11--CP15) and add adversarial continuity and BTANT-bridge checkpoints (CP16--CP20) that bring v1.0's $6/7$ coverage to $7/7$.

\bibliographystyle{plainnat}
\bibliography{references}

@article{packer2023memgpt,
  title={{MemGPT}: Towards {LLMs} as Operating Systems},
  author={Packer, Charles and Wooders, Sarah and Lin, Kevin and Fang, Vivian and Patil, Shishir G. and Stoica, Ion and Gonzalez, Joseph E.},
  journal={arXiv preprint arXiv:2310.08560},
  year={2023}
}

@article{chhikara2025mem0,
  title={Mem0: Building Production-Ready {AI} Agents with Scalable Long-Term Memory},
  author={Chhikara, Prateek and Khant, Dev and Aryan, Saket and Singh, Taranjeet and Yadav, Deshraj},
  journal={arXiv preprint arXiv:2504.19413},
  year={2025}
}

@article{ai2025memorybench,
  title={{MemoryBench}: A Benchmark for Memory and Continual Learning in {LLM} Systems},
  author={Ai, Qingyao and Tang, Yichen and Wang, Changyue and Long, Jianming and Su, Weihang and Liu, Yiqun},
  journal={arXiv preprint arXiv:2510.17281},
  year={2025}
}

@article{tavakoli2025beam,
  title={Beyond a Million Tokens: Benchmarking and Enhancing Long-Term Memory in {LLMs}},
  author={Tavakoli, Mohammad and Salemi, Alireza and Ye, Carrie and Abdalla, Mohamed and Zamani, Hamed and Mitchell, J. Ross},
  journal={arXiv preprint arXiv:2510.27246},
  year={2025}
}

@article{tanguturi2026atant,
  title={{ATANT}: An Evaluation Framework for {AI} Continuity},
  author={Tanguturi, Samuel Sameer},
  journal={arXiv preprint arXiv:2604.06710},
  year={2026}
}

@article{maharana2024locomo,
  title={Evaluating Very Long-Term Conversational Memory of {LLM} Agents},
  author={Maharana, Adyasha and Lee, Dong-Ho and Tulyakov, Sergey and Bansal, Mohit and Barbieri, Francesco and Fang, Yuwei},
  journal={Proceedings of ACL 2024},
  year={2024}
}

@article{wu2024longmemeval,
  title={{LongMemEval}: Benchmarking Chat Assistants on Long-Term Interactive Memory},
  author={Wu, Di and Wang, Hongwei and Yu, Wenhao and Zhang, Yunsheng and Chang, Kai-Wei and Yu, Dong},
  journal={Proceedings of ICLR 2025},
  year={2025}
}

@article{rasmussen2025zep,
  title={Zep: A Temporal Knowledge Graph Architecture for Agent Memory},
  author={Rasmussen, Preston and Paliychuk, Pavlo and Beauvais, Travis and Ryan, Jack and Chalef, Daniel},
  journal={arXiv preprint arXiv:2501.13956},
  year={2025}
}

@article{packer2024letta,
  title={Letta: Stateful Agents with Persistent Memory and Tool Use},
  author={Packer, Charles and Wooders, Sarah and Lin, Kevin and others},
  journal={Letta Technical Report},
  year={2024}
}

@article{hsieh2024ruler,
  title={{RULER}: What's the Real Context Size of Your Long-Context Language Models?},
  author={Hsieh, Cheng-Ping and Sun, Simeng and Kriman, Samuel and Acharya, Shantanu and Rekesh, Dima and Jia, Fei and Zhang, Yang and Ginsburg, Boris},
  journal={Proceedings of COLM 2024},
  year={2024}
}

\appendix

\section{Per-cell Justification for Table~\ref{tab:matrix}}
\label{app:cells}

This appendix provides per-cell reasoning for the property-coverage assignments in Table~\ref{tab:matrix}. For each (benchmark, property) cell we state the benchmark's operation on the property's evaluation surface and the rubric assignment that follows. The appendix is defensive armor: it exists to make individual cell challenges addressable without re-opening the aggregate argument.

\subsection{LOCOMO (row 1)}
\textbf{Pers $\times$.} LOCOMO is evaluated single-shot: the system ingests the full conversation transcript and answers all questions in one pass. There is no process-termination test between ingest and query; persistence as v1.0 defines it cannot be measured.

\textbf{Upd $\times$.} No item in the 1{,}986-question corpus presents a fact, then an update, then asks both for the current state and for the prior state. Supersession is not in the test surface.

\textbf{Tmp $\circledcirc$.} Category 2 (321 items, 77\% beginning with ``when'') tests date and duration resolution from transcript text. This is a subset of temporal ordering but does not test current-vs-past status or supersession ordering, both of which v1.0 Property 3 requires.

\textbf{Dis $\circledcirc$.} Multi-speaker conversations introduce entity overlap incidentally (two people named X, two events of type Y) but the benchmark does not systematically probe disambiguation. Partial credit.

\textbf{Rec $\times$.} Category 4 (841 items) gold answers are paraphrases of narrator utterances, not situation reconstructions. A system returning the exact substring that appears in the transcript passes without reconstructing anything.

\textbf{MdI $\times$.} Writer and reader are the same model in the intended evaluation setup.

\textbf{Ops $\times$.} Single domain (personal dialogue). No institutional or cross-surface test.

\subsection{LongMemEval (row 2)}
\textbf{Pers $\times$.} Single-prompt evaluation. No process termination.

\textbf{Upd $\times$.} The ``knowledge updates'' category tests within-prompt override, not cross-session supersession. Under strict v1.0 definition, this is $\times$.

\textbf{Tmp $\times$.} Temporal reasoning category exists but operates within-context; current-vs-past status is not an evaluation axis.

\textbf{Dis $\times$.} No systematic entity-disambiguation probe.

\textbf{Rec $\times$.} Answers are direct retrieval, not state reconstruction.

\textbf{MdI $\times$.} Same model for ingestion and query.

\textbf{Ops $\times$.} Chat-assistant-shaped; single surface.

\subsection{BEAM (row 3)}
\textbf{Pers $\times$.} Long-context single-pass.

\textbf{Upd $\times$.} Not in the evaluation surface.

\textbf{Tmp $\times$.} Extended reasoning tasks include some temporal-adjacent items but operate within one context.

\textbf{Dis $\times$.} Not tested.

\textbf{Rec $\times$.} Retrieval-plus-light-reasoning; not reconstruction.

\textbf{MdI $\times$.} Same model throughout.

\textbf{Ops $\times$.} Synthetic long-context; single surface.

\subsection{MemoryBench (row 4)}
\textbf{Pers $\circledcirc$.} Storage-and-retrieval is tested in isolation; the storage medium persists to disk, but the benchmark does not include an explicit process-termination round-trip test. Partial credit.

\textbf{Upd $\times$.} Not scored as a property.

\textbf{Tmp $\times$.} Not scored.

\textbf{Dis $\times$.} Not systematically tested.

\textbf{Rec $\times$.} Retrieval-over-chunks; not reconstruction.

\textbf{MdI $\times$.} Not tested.

\textbf{Ops $\times$.} Single-surface benchmark.

\subsection{Zep Eval (row 5)}
\textbf{Pers $\circledcirc$.} Graph storage persists in production; the benchmark does not isolate this as a scored axis.

\textbf{Upd $\times$.} Temporal edges exist in the graph; not scored as supersession behavior.

\textbf{Tmp $\times$.} Temporal annotations exist in the data model; current-vs-past distinction is not a scored metric.

\textbf{Dis $\circledcirc$.} Entity resolution is architecturally supported and partially probed in multi-entity retrieval tasks.

\textbf{Rec $\times$.} No situation-synthesis scored task.

\textbf{MdI $\times$.} Not tested.

\textbf{Ops $\times$.} Agent-memory-shaped; single surface in benchmark.

\subsection{MemGPT / Letta (row 6)}
\textbf{Pers $\times$ (as measured).} Tools for persistent memory exist; the benchmark scores whether they are called, not whether their contents persist correctly across terminations.

\textbf{Upd $\times$.} Not scored.

\textbf{Tmp $\times$.} Not scored.

\textbf{Dis $\times$.} Not scored.

\textbf{Rec $\times$.} Not scored.

\textbf{MdI $\times$.} Not tested.

\textbf{Ops $\circledcirc$.} Multi-task agentic evaluations exercise cross-surface behavior to a limited degree.

\subsection{RULER (row 7)}
\textbf{All properties $\times$.} RULER is a context-window stress test; every property in v1.0 is outside its measurement surface by design.

\subsection{ATANT v1.0 (row 8)}
\textbf{Pers $\checkmark$.} CP2 + process-termination protocol (v1.0 \S4, Principle 5) tests write-then-restart-then-read.

\textbf{Upd $\checkmark$.} CP9 + cumulative-mode supersession tests present contradictory facts and score current-state retrieval.

\textbf{Tmp $\checkmark$.} CP9 scores temporal resolution, sequencing, and current-vs-past status.

\textbf{Dis $\checkmark$.} Cumulative mode (50 and 250 stories coexisting) tests disambiguation under memory load. This is v1.0's distinctive axis.

\textbf{Rec $\circledcirc$.} CP7 tests convergence (multi-trace retrieval); CP8 scores the final answer's keyword set, not its narrative coherence \citep{tanguturi2026atant}. Situation-synthesis is partially covered but not graded for internal consistency; v2.0 addresses this.

\textbf{MdI $\checkmark$.} Principle 1 (v1.0 \S4) mandates LLM-free evaluation; writer and reader are the structural pipeline, not a model.

\textbf{Ops $\circledcirc$.} Corpus spans 6 life domains; the operational-across-surfaces property is supported structurally but the benchmark does not include institutional or multi-application deployment tests. Partial credit.

\subsection{ATANT v2.0 Proposed (row 9)}
$7/7$ is the design target. CP11--CP15 (situation synthesis, what-changed, multi-trace fusion, narrative coherence, emotional arc) close Rec to $\checkmark$. Extended corpus across 2+ operational domains (clinic / library stories in addition to personal) closes Ops to $\checkmark$. Specification appears in the v2.0 extension paper.

\section{Proposed Fix for LOCOMO \texttt{answer\_matches}}
\label{app:fix}

The empty-gold matcher defect documented in \S\ref{sec:locomo} is correctable with a five-line change. We sketch the patch for upstream submission:

\begin{verbatim}
def answer_matches(predicted, gold):
    gold_clean = str(gold or "").strip()
    pred_clean = str(predicted or "").strip()

    # Empty-gold refusal: the system should refuse or return nothing
    if not gold_clean:
        if not pred_clean:
            return True
        refusal_markers = (
            "i don't know", "i do not know", "no information",
            "insufficient", "cannot determine", "unknown"
        )
        return any(m in pred_clean.lower() for m in refusal_markers)

    if not pred_clean:
        return False
    # ... existing substring match logic ...
\end{verbatim}

With this fix, category-5 questions become scorable. A system that refuses on empty gold is credited; a system that fabricates is not. The fix does not alter scoring of categories 1--4.

Applying the fix raises Kenotic's LOCOMO score in proportion to its refusal discipline on category 5. We do not report a post-fix number because (a) the fix has not been accepted upstream at time of writing, and (b) the structural critique of categories 1--4 (substring matching on paraphrase) is independent of the fix.

\end{document}